\documentclass[10pt,twocolumn,letterpaper]{article}

\usepackage[pagenumbers]{cvpr} 

\usepackage{graphicx}
\usepackage{amsmath}
\usepackage{amssymb}
\usepackage{booktabs}
\usepackage{multirow}

\usepackage[pagebackref,breaklinks,colorlinks, citecolor={blue}]{hyperref}

\usepackage[capitalize]{cleveref}
\crefname{section}{Sec.}{Secs.}
\Crefname{section}{Section}{Sections}
\Crefname{table}{Table}{Tables}
\crefname{table}{Tab.}{Tabs.}

\def\cvprPaperID{1581} 
\def\confName{CVPR}
\def\confYear{2023}
\newcommand{\mypar}[1]{\vspace{-4mm}\paragraph{#1}}

\begin{document}

\title{Boosting Detection in Crowd Analysis via Underutilized Output Features}

\author{Shaokai Wu\textsuperscript{1*} 
\quad
Fengyu Yang\textsuperscript{2*} \vspace{3mm} \\
Jilin University\textsuperscript{1} \quad University of Michigan\textsuperscript{2} \vspace{1mm} \\
}
\maketitle
{\let\thefootnote\relax\footnotetext{{\textsuperscript{*} Indicates equal contribution.}}}
\def\projecturl{https://fredfyyang.github.io/Crowd-Hat/}

\begin{abstract}
    Detection-based methods have been viewed unfavorably in crowd analysis due to their poor performance in dense crowds. However, we argue that the potential of these methods has been underestimated, as they offer crucial information for crowd analysis that is often ignored. Specifically, the area size and confidence score of output proposals and bounding boxes provide insight into the scale and density of the crowd. To leverage these underutilized features, we propose Crowd Hat, a plug-and-play module that can be easily integrated with existing detection models. This module uses a mixed 2D-1D compression technique to refine the output features and obtain the spatial and numerical distribution of crowd-specific information. Based on these features, we further propose region-adaptive NMS thresholds and a decouple-then-align paradigm that address the major limitations of detection-based methods. Our extensive evaluations on various crowd analysis tasks, including crowd counting, localization, and detection, demonstrate the effectiveness of utilizing output features and the potential of detection-based methods in crowd analysis. Project Page: \url{https://fredfyyang.github.io/Crowd-Hat/}

\end{abstract}

\section{Introduction}
Crowd analysis is one of a critical area in computer vision~\cite{Wu_2022_CVPR, yang2022touch, ji2019human, li2022invariant, chen2023deepmapping2, Yang_2022_CVPR, What_Transferred_Dong_CVPR2020, zhao2022rbc, MaQBL23, Li2023kdd} due to its close relation with humans and its wide range of applications in public security, resource scheduling, crowd monitoring \cite{NWPU, SUA-Fully, AutoScale}. This field can be divided into three concrete tasks: crowd counting \cite{MAN, dm, bys}, crowd localization \cite{p2p, topo, GL}, and crowd detection \cite{lsc, PSDNN, SDNet}. While most existing methods mainly focus on the first two tasks due to the extreme difficulty of detecting dense crowds, simply providing the number of the crowd or representing each person with a point is insufficient for the growing real-world demand. Crowd detection, which involves localizing each person with a bounding box, supports more downstream tasks, such as crowd tracking \cite{tracking} and face recognition \cite{face-recognition}. Therefore, constructing a comprehensive crowd analysis framework that addresses all three tasks is essential to meet real-world demands.
    
\begin{figure}[t]
  \centering
  \setlength{\belowcaptionskip}{-0.6cm}
  \includegraphics[width=0.47\textwidth]{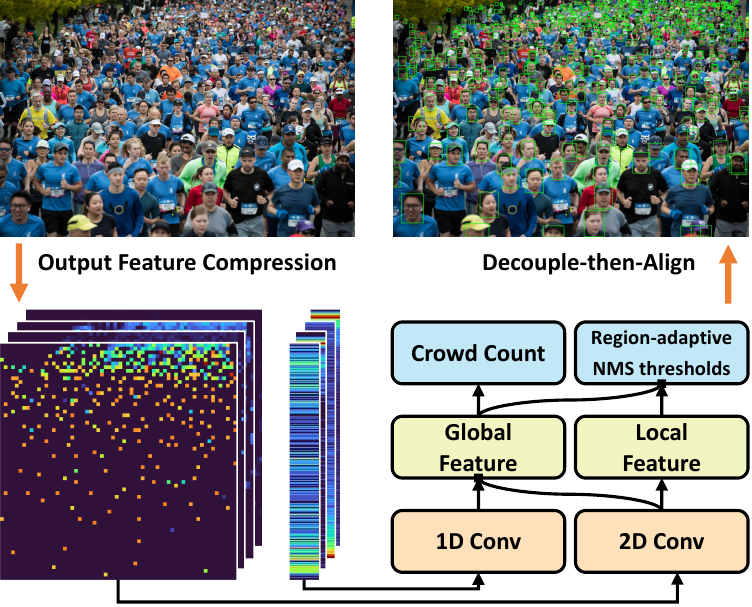}
  \caption{Our approach involves extracting output features from the detection outputs and refining them into 2D compressed matrices and 1D distribution vectors. These features are then encoded into local and global feature vectors to regress region-adaptive NMS thresholds and the crowd count. We select the final output bounding boxes using the decouple-then-align paradigm.}
  \label{fig:first}
\end{figure}

Although object detection may seem to meet the demands above, most relevant research views it pessimistically, especially in dense crowds \cite{MCNN, dm, MAN, p2p, NWPU}. First and foremost, compared to general object detection scenarios with bounding box annotations, most crowd analysis datasets only provide limited supervision in the form of point annotations. As a result, detection methods are restricted to using pseudo bounding boxes generated from point labels for training \cite{lsc, PSDNN, SDNet}. However, the inferior quality of these pseudo bounding boxes makes it difficult for neural networks to obtain effective supervision \cite{p2p, topo}.

Secondly, the density of crowds varies widely among images, ranging from zero to tens of thousands \cite{lsc, NWPU, SH-AB, ucf-qnrf, jhu}, and may vary across different regions in the same image, presenting a significant challenge in choosing the allowed overlapping region in Non-Maximum-Suppression (NMS). A fixed NMS threshold is often considered a hyperparameter, but it yields a large number of false positives within low-density crowds and false negatives within high-density crowds, as criticized in \cite{p2p}.

Thirdly, current detection methods adopt the detection-counting paradigm \cite{bys, dm} for crowd counting, where the number of humans is counted by the bounding boxes obtained from the detection output. However, the crowd detection task is extremely challenging without bounding box labels for training, resulting in a large number of mislabeled and unrecognized boxes in crowds \cite{p2p, NWPU}. This inaccurate detection result makes the detection-counting paradigm perform poorly, yielding inferior counting results compared to density-based methods.

\label{area-confidence}
In this paper, we ask: \emph{Has the potential of object detection been fully discovered?} {Our findings suggest that crucial information from detection outputs, such as the size and confidence score of proposals and bounding boxes, are largely disregarded. This information can provide valuable insights into crowd-specific characteristics. For instance, in dense crowds, bounding boxes tend to be smaller with lower confidence scores due to occlusion, while sparse crowds tend to produce boxes with higher confidence scores.}
    
To this end, we propose the a module on top of the Head of detection pipeline to leverage these underutilized detection outputs. We name this module as \textbf{``Crowd Hat"} because it can be adapted to different detection methods, just as a hat can be easily put on different heads. Specifically, we first introduce a mixed 2D-1D compression to refine both spatial and numerical distribution of output features from the detection pipeline. {We further propose a NMS decoder to learn region-adaptive NMS thresholds from these features, which effectively reduces false positives under low-density regions and false negatives under high-density regions. Additionally, we use a decouple-then-align paradigm to improve counting performance by regressing the crowd count from output features and using this predicted count to guide the bounding box selection. Our Crowd Hat module can be integrated into various one-stage and two-stage object detection methods, bringing significant performance improvements for crowd analysis tasks.} Extensive experiments on crowd counting, localization and detection demonstrate the effectiveness of our proposed approach.

Overall, the main contributions of our work can be summarized as the following:
    \vspace{-3mm}
    \begin{itemize}
    \setlength\itemsep{-1mm}
    \item To the best of our knowledge, we are the first to consider detection outputs as valuable features in crowd analysis and propose the mixed 2D-1D compression to refine crowd-specific features from them.
    \item We introduce region-adaptive NMS thresholds and a decouple-then-align paradigm to mitigate major drawbacks of detection-based methods in crowd analysis.
    \item We evaluate our method on public benchmarks of crowd counting, localization, and detection tasks, showing our method can be adapted to different detection methods while achieving better performance. 
    \end{itemize}

\section{Related Work}
\vspace{3mm}
\mypar{Density-Based Methods}
{Density-based methods have been continuously improved since first proposed in \cite{density-base}, for their superior performance and high efficiency in counting tasks. In this paradigm, a network is trained to map an input image to the crowd density map, thus the number of crowds is computed by summing the whole density map. 
While most advanced crowd counting methods are density-based \cite{dm,bys,MAN,density1,density2,density3,density4,density5,density6,density7,density8,density9,density10}, they tend to neglect spatial information \cite{topo,lsc,p2p}, resulting in poor performance in individual pinpointing and bounding box provision for crowd heads \cite{PSDNN,lsc,SDNet}.}

\mypar{Localization-Based Methods}
To address the shortcomings of density-based methods in localization, researchers have proposed localization-based methods such as \cite{topo,p2p,RAZ_LOC,ucf-qnrf,scalnet}. These methods achieve better performance in crowd localization and count crowds by summing the total number of points. However, while these methods outperform density-based methods in localization, their counting performance is generally worse \cite{MAN,CCTrans,SAA}, and they still fall short in meeting the needs of crowd detection \cite{lsc,SDNet,PSDNN}.

\mypar{Detection-Based Methods}
Although detection-based methods are capable of resolving detection, localization, and counting tasks simultaneously, current research shows a pessimistic view of this paradigm. The fundamental capability of detection typically requires a significant number of bounding box labels for training, which are often unavailable in many crowd datasets \cite{ucf-cc-50,ucf-qnrf,SH-AB}. Only a few methods, such as \cite{PSDNN,lsc,SDNet}, attempt to train a detection network with pseudo box labels generated from point annotations. However, these methods suffer from inaccurate bounding boxes due to a fixed NMS process, which leads to too many false positives. As a result, detection-based methods generally perform worse in counting and localization tasks.

\begin{figure*}[t]
  \centering
  \includegraphics[width=1\textwidth]{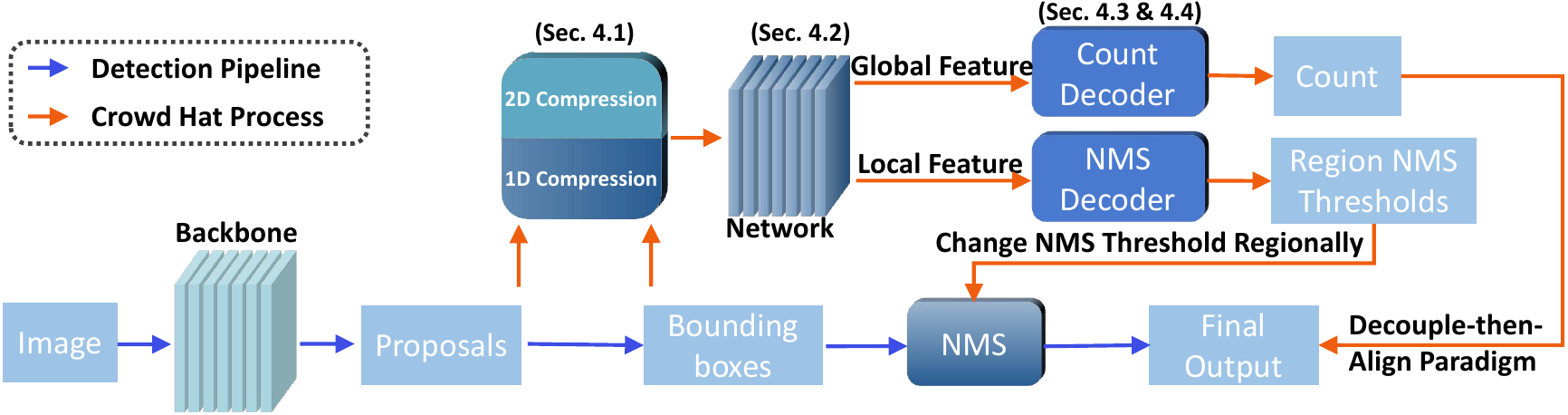}
  \caption{\textbf{Overview of the Crowd Hat module.} We adopt a two-stage detection method PSDNN as our detection pipeline for illustration.}
  \label{fig:overview}
   \vspace{-5mm}
\end{figure*} 


\section{Preliminary}
\label{sec:3}
 Given a set of pair $(I, A)$ in the dataset, we define $I \in \mathbb{R}^{3\times H \times W}$ as the input image containing $N$ people and $A = \{a_1,a_2,...,a_N\}$ as the list of corresponding point annotations where $a_k = (a_k^x,a_k^y)$ is the center location of $k$ th head. $B=\{b_1,b_2,...,b_k,...,b_n\}$ denotes the set of bounding boxes output by the network with $n$ predictions, and the $k$ th box is $b_k=(b_k^x,b_k^y,b_k^w,b_k^h,b_k^c)$ where $b_k^x$ and $b_k^y$ is the coordinates of the center point, $b_k^w,b_k^h$ is the width and height, and $b_k^c$ is the confidence of this box. For two-stage methods with region proposals, we denote $P=\{p_1,p_2,...,p_k,...,p_m\}$ as the set of proposals with $m$ predictions. Likewise the $k$ th proposal is $p_k=(p_k^x,p_k^y,p_k^w,p_k^h,p_k^c)$ with $p_k^x,p_k^y$ as center coordinates, $p_k^w,p_k^h$ as the width and height, and $p_k^c$ as the confidence of the proposal. Note that proposals and bounding boxes refer to the predictions before applying NMS and confidence score filtering.
 
 All methods in our paper were trained using only point annotations to ensure fair comparisons \cite{SH-AB,ucf-qnrf,NWPU}. Therefore, we generated pseudo bounding box labels from point annotations for training all detection methods, following common practice~\cite{lsc,SDNet,PSDNN}.


\section{Methodology: Crowd Hat}
\begin{figure*}[t]
  \centering
  \includegraphics[width=1\textwidth]{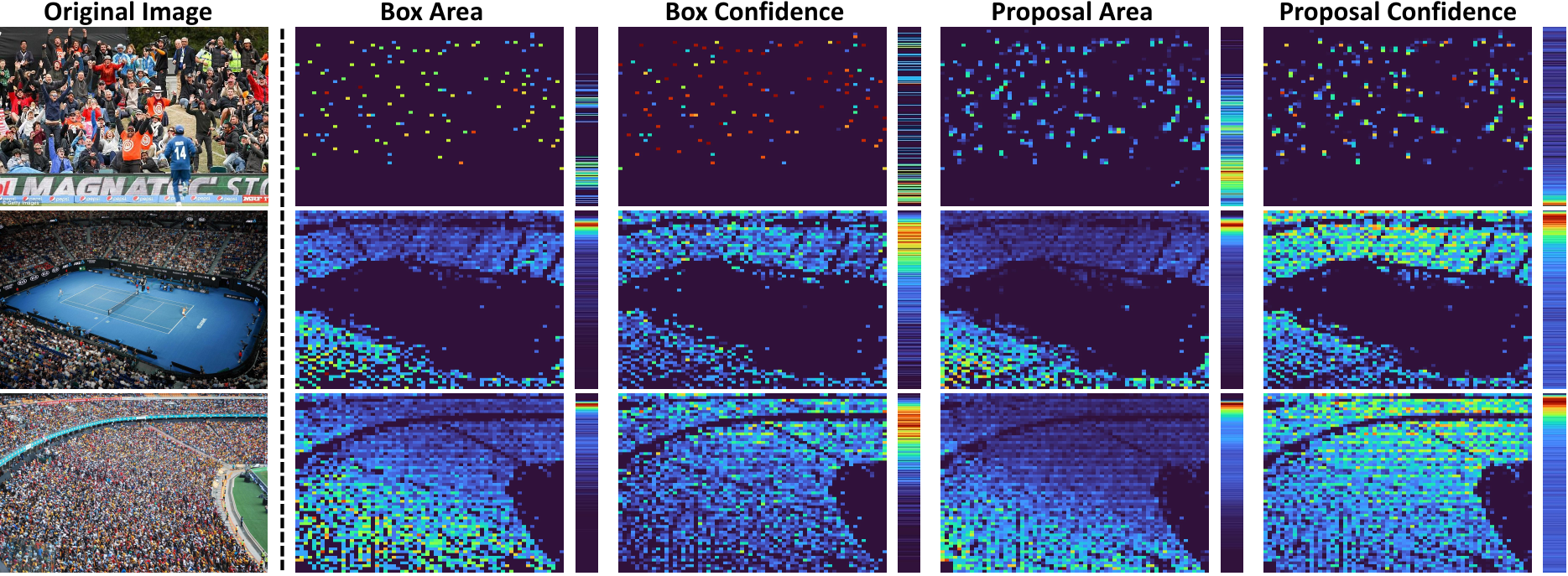}
  \caption{{\bf Visualization of output feature compression.} {We present the 2D compression matrices (left) and 1D distribution vectors (to the right) for each output feature. In the 1D distribution vectors, we use 0 to denote the top of the vector and 1 for the bottom. Additionally, we provide the original image in the leftmost column for reference.} \emph{Zoom in for better visualization.}}
  \label{fig:compression}
   \vspace{-5mm}
\end{figure*}  
In our paper, we define detection outputs as the predicted bounding boxes and proposals from the detection pipeline. We find these outputs convey abundant crowd-specific information, making them valuable assets for crowd analysis tasks. {In particular, we adopt two output features, namely ``area size" and ``confidence score" from detection outputs.  Compared to feature maps extracted from convolution layers (CNN features), output features focus mostly on humans, the foreground of the image, and are considered relatively ``pure" features for crowd analysis tasks. Therefore, we propose the Crowd Hat module to mine and utilize these output features, as shown in Figure \ref{fig:overview}. In this section, we use the two-stage detection method PSDNN \cite{PSDNN} as the detection pipeline, and other one-stage detection methods \cite{SDNet,lsc} can be easily adapted.}

\subsection{Output Feature Compression}
\label{sec:4.1}
The original format of detection outputs is a list of 5D vectors (see definitions in Section~\ref{sec:3}). Since the number of generated bounding boxes $n$ and the number of proposals $m$ vary among images, it is hard to pass these irregular vectors directly into neural networks. While mapping the output features directly back to the input image according to the center coordinates of detection outputs may seem like a trivial approach, the resulting feature maps will be too sparse to convey representative information since the number of predicted proposals or bounding boxes is far less than the number of pixels per image. To address this issue, we propose a mixed 2D-1D compression method to further refine the output features and obtain the spatial and numerical distribution of these crowd-specific information. We show visualization results of different features from the 2D-1D compression method in Figure \ref{fig:compression}.

\vspace{-3mm}
\subsubsection{2D Compression}
{To determine the spatial distribution of crowd density in an image, we propose 2D compression using a matrix $M\in R^{S\times S}$ to compress each output feature into patches. We map the proposal or bounding box to the input image based on its center coordinates, divide the image into $S\times S$ patches of equal height and width, and sum up each output feature located within each patch to obtain the corresponding element in the compression matrix $M$.}

{Consider compressing bounding box area using a compression matrix. To calculate the normalized area of the k-th bounding box, we multiply its width by the image height and divide the result by the product of the image's width and height. This normalization step removes the impact of different image resolutions. Specifically, the normalized size is calculated as  $\frac{b_k^w}{W} \cdot \frac{b_k^h}{H}$. We denote $M_B^A$ as the compressed matrix of bounding box area. Thus the formula yields:}

\begin{flalign}
  M_B^A(i,j)=\sum_{k=1}^{n}[\left \lfloor \frac{b_k^x}{w_0} \right \rfloor =i]\cdot 
[\left \lfloor \frac{b_k^y}{h_0} \right \rfloor =j] \cdot \frac{b_k^w}{W} \cdot \frac{b_k^h}{H}  && 
\label{feature:MBA}
\end{flalign}
where two indicator functions $[\left \lfloor \frac{b_k^x}{w_0} \right \rfloor =i]$ and $[\left \lfloor \frac{b_k^y}{h_0} \right \rfloor =j]$ are equal to 1 only if the bounding box $b_k$ belongs to the patch indexed $(i,j)$ and 0 otherwise.

{We use the notation $M_B^C$ to represent the compressed matrix of bounding box confidence scores, and $M_P^A$ and $M_P^C$ to represent the compressed matrices of proposal area size and confidence score, respectively. If a one-stage detection method does not generate proposals, we only use $M_B^A$ and $M_B^C$ as the compressed matrices. These matrices are formally defined as follows:}

\begin{align}
    M_B^C(i,j)=\sum_{k=1}^{n}[\left \lfloor \frac{b_k^x}{w_0} \right \rfloor =i]\cdot 
[\left \lfloor \frac{b_k^y}{h_0} \right \rfloor =j] \cdot \sigma(b_k^c) &&
\label{feature:MBC}
\end{align}

\begin{flalign}
  M_P^A(i,j)=\sum_{k=1}^{m}[\left \lfloor \frac{p_k^x}{w_0} \right \rfloor =i]\cdot 
[\left \lfloor \frac{p_k^y}{h_0} \right \rfloor =j] \cdot \frac{p_k^w}{W} \cdot \frac{p_k^h}{H} &&
\label{feature:MPA}
\end{flalign}

\begin{align}
    M_P^C(i,j)=\sum_{k=1}^{m}[\left \lfloor \frac{p_k^x}{w_0} \right \rfloor =i]\cdot 
[\left \lfloor \frac{p_k^y}{h_0} \right \rfloor =j] \cdot \sigma(p_k^c) &&
\label{feature:MPC}
\end{align}

\subsubsection{1D Compression}
\label{sec:1d-compression}
\begin{figure*}[t]
  \centering
  \setlength{\belowcaptionskip}{-0.5cm}
  \includegraphics[width=1\textwidth]{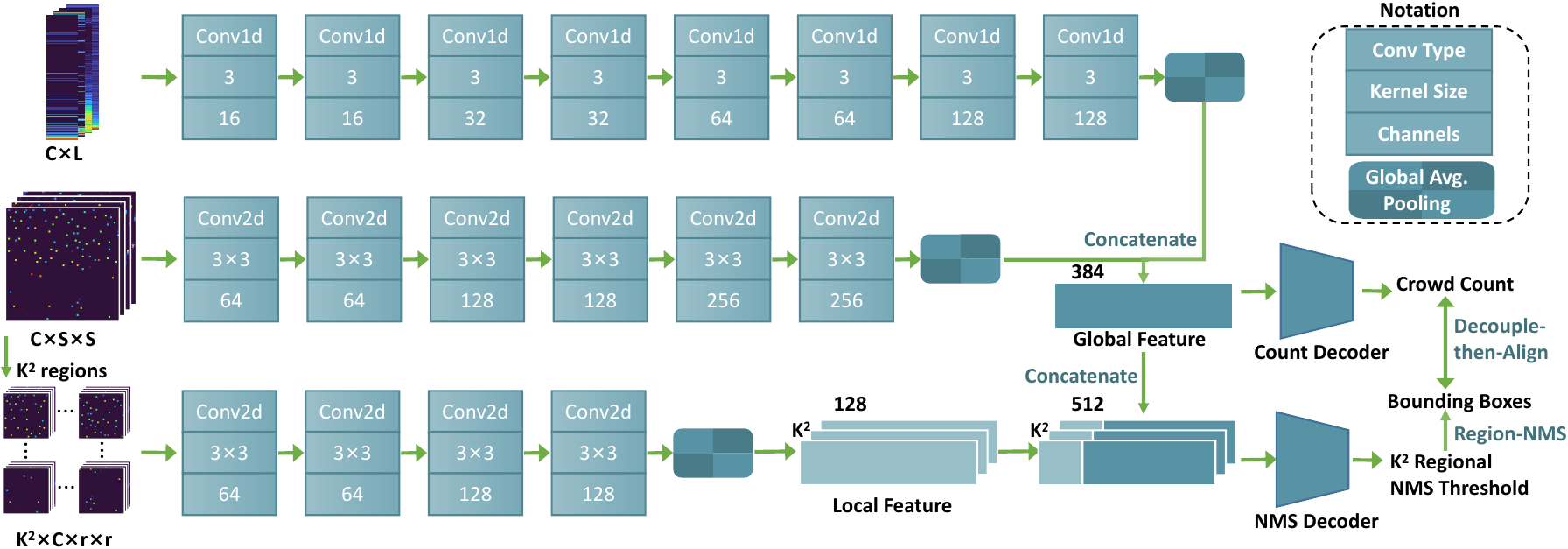}
  \caption{{\bf Detailed neural network structure of Crowd Hat module}. The 2D compressed matrices and 1D distribution vectors are further transformed into global and local feature vectors via convolutional neural networks. {Each convolution layer is followed by ReLU activation and a max-pooling layer, except for the last layer where we use global average pooling.}}
  \label{fig:convolution}
\end{figure*}
Crowd density varies greatly within and among images, with some images having densities ranging from zero to tens of thousands~\cite{lsc,NWPU,SH-AB,ucf-qnrf,jhu}. To determine the overall crowd density of an image, we propose a 1D compression method that finds the numerical distribution of output features within the image. For instance, a low overall distribution of output bounding box area sizes could indicate a high crowd density in the scene.

Our proposed 1D compression method works as follows: first, we normalize the confidence score and area size values to a range of 0 to 1. Next, we divide this range into $L$ discrete intervals, where the $i$-th interval is $[\frac{i}{L},\frac{i + 1}{L})$. We then calculate the number of values that fall into each interval to form a histogram and obtain the numerical distribution where the $i$-th value of the histogram represents the number of values that fall into the interval $[\frac{i}{L},\frac{i + 1}{L})$.

{To normalize the output features into the interval of 0 to 1, we use a two-step process. First, we multiply the output feature by a scaling coefficient $\alpha \geq 1$, which is a hyperparameter. This step is necessary to ensure that the distribution of output features is distinguishable after nonlinear mapping. The raw values of area size and confidence score may be numerically congested, causing them to fall into the same interval or nearby intervals after nonlinear mapping. Second, we apply a nonlinear mapping function to the output feature to limit its range to [0, 1]. Note that area size is always greater than zero, while confidence can be either positive or negative, thus we use the sigmoid function $\sigma(x)$ for the confidence score and the hyperbolic tangent function $tanh(x)$ for the area size. }

{For instance, we transform the confidence score of the $k$-th bounding box, $b_k^c$, to $\sigma(b_k^c\cdot \alpha)$. This value falls into the interval $\sigma(b_k^c\cdot \alpha)/(\frac{1}{L}) = \sigma(b_k^c\cdot \alpha) \cdot L$. We refer to the distribution vector of the bounding box confidence score as $V_B^C$ and its scaling coefficient as $\alpha_B^C$.} The formula yields:
\vspace{-2mm}
\begin{align}
    V_B^C(i)=\sum_{k=1}^{n} [\left \lfloor \sigma(b_k^c\cdot \alpha_B^C) \cdot L \right \rfloor =i] &&
\label{feature:VBC}
\end{align}

{Similarly, we denote $V_B^A$ as the distribution vector of the bounding box area size and $V_P^A,V_P^C$ as those of proposal area size and confidence score. The scaling coefficients for the corresponding output features are denoted as $\alpha_B^A$, $\alpha_P^A$, and $\alpha_P^C$, respectively. For one-stage detection methods that do not generate proposals, we only use $V_B^A$ and $V_B^C$ from the predicted bounding boxes. In formal terms:}
\begin{align}
    V_B^A(i)=\sum_{k=1}^{n} [\left \lfloor tanh(\frac{b_k^w}{W}\cdot \frac{b_k^h}{H}\cdot \alpha_B^A) \cdot L \right \rfloor =i] &&
\label{feature:VBA}
\end{align}

\begin{align}
    V_P^A(i)=\sum_{k=1}^{m} [\left \lfloor tanh(\frac{p_k^w}{W}\cdot \frac{p_k^h}{H}\cdot \alpha_P^A)\cdot L \right \rfloor =i] &&
\label{feature:VPA}
\end{align}

\begin{align}
    V_P^C(i)=\sum_{k=1}^{m} [\left \lfloor \sigma(p_k^c\cdot \alpha_P^C)\cdot L \right \rfloor =i] &&
\label{feature:VPC}
\end{align}

\subsection{Crowd Hat Network}
\label{sec:4.2}
{To aggregate information from the different output features above, we stack the 2D compressed matrices to form a tensor $t_{2d} \in R^{C \times S \times S}$, and the distribution vectors from 1D compression are stacked to form a tensor $t_{1d} \in R^{C \times L}$. Here, $C$ is the number of output features used, with $C=4$ for two-stage methods and $C=2$ for one-stage methods. These tensors are then passed into our Crowd Hat network to obtain global and local features, as described below. The detailed structure is shown in Figure~\ref{fig:convolution}.
}

\vspace{0.2cm}
\noindent {\bf Global Feature}
\label{sec:4.2.1}
{To incorporate both the spatial information from $t_{2d}$ and the numerical distribution information from $t_{1d}$, we use 2D convolutions to further encode $t_{2d}$ and 1D convolutions for $t_{1d}$, as shown in Figure \ref{fig:convolution}. After global average pooling, we concatenate both of them to form the global feature vector $F_g$.}

\vspace{0.2cm}
\noindent {\bf Local Feature}
\label{split}
{To capture the high variation of crowd density within an image and support our region-adaptive NMS, we introduce local features by dividing $t_{2d}$ into fixed-sized patches and encoding them using neural networks. We split $t_{2d}$ into $K\times K$ patches and then pass them through a 2D convolutional neural network with global average pooling to generate local feature vectors $[F_l^1, F_l^2, ... F_l^{K^2}]$.
}

\begin{table*}
\footnotesize
\setlength{\abovecaptionskip}{0.15cm}
\centering
\caption{\textbf{Quantiative comparisons in crowd counting}, best in {\bf bold}, second in \underline{underline}. All results are from corresponding papers or official implementations, and official records from NWPU-Crowd benchmark.}
\resizebox{\textwidth}{!}{
\begin{tabular}{lccccccccccc}
\toprule
\multicolumn{1}{c}{\multirow{2}{*}{\textbf{Method}}} &
\multicolumn{1}{c}{\multirow{2}{*}{\textbf{Type}}} &
\multicolumn{2}{c}{\textbf{ShanghaiTech A}} &
\multicolumn{2}{c}{\textbf{ShanghaiTech B}} &
\multicolumn{2}{c}{\textbf{JHU-Crowd++}} & \multicolumn{2}{c}{\textbf{UCF-QNRF}} & \multicolumn{2}{c}{\textbf{NWPU-Crowd}} \\ 
\cmidrule(lr){3-4} \cmidrule(lr){5-6} \cmidrule(lr){7-8} \cmidrule(lr){9-10} \cmidrule(lr){11-12} & &
\multicolumn{1}{l}{MAE $\downarrow$} & \multicolumn{1}{l}{RMSE $\downarrow$} &
\multicolumn{1}{l}{MAE $\downarrow$} & \multicolumn{1}{l}{RMSE $\downarrow$} &
\multicolumn{1}{l}{MAE $\downarrow$} & \multicolumn{1}{l}{RMSE $\downarrow$} & \multicolumn{1}{l}{MAE $\downarrow$} & \multicolumn{1}{l}{RMSE $\downarrow$} & \multicolumn{1}{l}{MAE $\downarrow$} & \multicolumn{1}{l}{RMSE $\downarrow$}\\ \midrule
Topo-Count\cite{topo} (AAAI 2020) &\multirow{4}{*}{Localization-Based} &  61.2 & 104.6 & 7.8 & 13.7 & 60.9 & 267.4 &  89.0 & 159.0 & 107.8 & 438.5 \\
P2P-Net~\cite{p2p} (ICCV 2021) &  & \underline{52.7} & \underline{85.1} & 6.3 & 9.9 & 56.3 & 268.3 & 85.3 & 154.5 &  77.4 & 362.0 \\
GL~\cite{GL} (CVPR 2021) & & 61.3 & 95.4 & 7.3 & 11.7 & 59.9 & 259.5 & 84.3 & 147.5 & 79.3 & 346.1 \\
CLTR\cite{liang2022end} (ECCV 2022) & &  56.9 & 95.2 & 6.5 & 10.6 & 59.5 & 240.6 &  85.8 & 141.3 & 74.3 & 333.8 \\
\midrule
ADSCNet~\cite{ADSC} (CVPR 2020)& \multirow{4}{*}{Density-Based} & 55.4 & 97.7 & 6.4 & 11.3 & - & - & \textbf{71.3} & 132.5 & - & - \\
SUA-Fully \cite{SUA-Fully} (ICCV 2021)& & 66.9 & 125.6 & 12.3 & 17.9 & 80.1 & 305.3 & 119.2 & 213.3 & 111.7 & 443.2 \\
MAN~\cite{MAN} (CVPR 2022)& & 56.8 & 90.3 & - & -  & \underline{53.4} & \textbf{209.9} & 77.3 & \underline{131.5} &  76.5 & 323.0 \\ 
CrowdFormer~\cite{CrowdFormer} (IJCAI 2022)& & 56.9 & 97.4 & \textbf{5.7} & \underline{9.6} & - & - & 78.8 & 136.1 & \textbf{67.1} & \underline{301.6} \\ 
\midrule
LSC-CNN~\cite{lsc} (TPAMI 2021)& \multirow{3}{*}{Detection-Based} &  66.4 & 117.0 & 8.1 & 12.7 & 87.3 & 309.0 & 120.5 & 218.2 &115.4 & 418.5 \\
SDNet \cite{SDNet} (TIP 2021)& &  65.1 & 104.4 & 7.8 & 12.6 & 78.8 & 295.4 &  102.1 & 176.0 & 100.2 & 385.8 \\
PSDNN~\cite{PSDNN} (CVPR 2019)&  & 70.2 & 125.8 & 9.1 & 14.2 & 95.7 & 344.3 & 137.5 & 240.1 & 140.7 & 553.6  \\
\midrule
LSC-CNN + Crowd Hat (ours)& \multirow{3}{*}{Detection-Based}& 60.2 & 95.7 & 7.1 & 11.3 &  63.0 & 270.9 &  84.7 & 150.2 & 90.6 & 336.6 \\
SDNet + Crowd Hat (ours)& & 53.4 & 87.2 & 6.5 & 10.4 &  56.9 & 251.5 &  81.0 & 139.4 & 73.7 & 321.0 \\
PSDNN + Crowd Hat (ours)& & \textbf{51.2} & \textbf{81.9} & \textbf{5.7} & \textbf{9.4} &  \textbf{52.3} & \underline{211.8} &  \underline{75.1} & \textbf{126.7} & \underline{68.7} & \textbf{296.9} \\ \bottomrule
\end{tabular}}
\label{tab:counting}
\vspace{-5mm}
\end{table*}

\subsection{Region-Adaptive NMS Decoder}
\label{NMS}
{We propose a NMS Decoder to address the challenge of varying crowd densities across regions. The region-adaptive NMS approach learns optimal NMS thresholds for each region, maximizing F1 score with current pseudo bounding box labels. To determine the pseudo NMS threshold labels for each region $[T_1, T_2, ..., T_{K^2}]$, we use a linear search algorithm, measuring model performance under different NMS thresholds ranging from 0 to 1 at a fixed step of $s$, and selecting the NMS threshold that leads to the highest F1 score for each region.}

{To train our NMS Decoder, we concatenate local and global features and pass them through an MLP, $P_N$, to generate region-adaptive NMS thresholds. We directly regress pseudo NMS threshold labels for training. The region NMS loss can be defined as follows:}
\begin{equation}
  \mathcal{L}_{nms} = \frac{1}{K^2}\sum_{i = 1}^{K^2}  |P_N(F_l^i\odot  F_g) - T_i|
\end{equation}
where $\odot $ denotes concatenation.

{During inference, we apply the learned pseudo NMS threshold labels from the NMS Decoder to perform region-adaptive NMS. For each region, we use the corresponding pseudo NMS threshold label as the NMS threshold to filter out redundant bounding boxes. Since different regions may have different crowd densities, the region-adaptive NMS can filter out more redundant bounding boxes in regions with high crowd density and retain more bounding boxes in regions with low crowd density, leading to better performance.}

\subsection{Decouple-then-Align Paradigm}
\label{sec:4.4}
{Detection-based methods for crowd counting suffer from the drawback of utilizing a detection-counting paradigm \cite{lsc,PSDNN,SDNet}, where crowd count is predicted by counting the number of output bounding boxes. This paradigm leads to entanglement between counting performance and detection and localization results, which is particularly problematic given that datasets typically provide only point annotations, resulting in limited supervision to train the detection pipeline.}

{To address this issue, we propose to decouple the detection and counting process by directly regressing the crowd count using global features $F_g$. Unlike some early methods that use CNN features for count regression, the output features we use provide valuable crowd-specific information, making them more suitable for direct count regression. We use a separate MLP, called the Count Decoder $P_C$, to predict the crowd count $\hat{n} = P_C(F_g)$, which is supervised by the ground truth crowd count obtained from point annotations. The loss function is formulated as follows:}

\begin{equation}
  \mathcal{L}_{count} = |P_C(F_g) - N|
\end{equation}

{However, using a separate count regression may cause confusion due to inconsistent results between the regression and detection process. The number of bounding boxes generated after NMS filtering $n_c$ may differ from the count output from the Count Decoder $\hat{n}$, leading to uncertainty about which number to reference. To address this issue, we prioritize the accuracy and reliability of the Count Decoder and select the min($\hat{n}$,$n_c$) bounding boxes with the highest confidences as the final results.}

\section{Experiments}
\begin{table*}[t]
\footnotesize
\setlength{\abovecaptionskip}{0.15cm}
\setlength{\belowcaptionskip}{-0.4cm}
\centering
\caption{\textbf{Comparisons in crowd localization}, best in {\bf bold}, second in \underline{underline}. All other results except for ours are from corresponding papers or official implementations, and official records from NWPU-Crowd benchmark.}
\resizebox{\textwidth}{!}{
\begin{tabular}{@{}lcccccccccccc@{}}
\toprule
\multirow{2}{*}{\textbf{Method}} & 
\multicolumn{1}{c}{\multirow{2}{*}{\textbf{Type}}}&
\multicolumn{3}{c}{\textbf{JHU-Crowd++}} &  & \multicolumn{3}{c}{\textbf{UCF-QNRF}} &  & \multicolumn{3}{c}{\textbf{NWPU-Crowd}} \\ \cmidrule(lr){3-5} \cmidrule(lr){7-9} \cmidrule(l){11-13} 
 &  & F1-m $\uparrow$ & Pre $\uparrow$ & Rec $\uparrow$ &  & F1-m $\uparrow$ & Pre $\uparrow$ & Rec $\uparrow$ &  & F1-m $\uparrow$ & Pre $\uparrow$ & Rec $\uparrow$ \\ \midrule
Topo-Count~\cite{topo} (AAAI 2020) &\multirow{5}{*}{Localization-Based}  & 57.6 & 62.6 & 53.4 &  &  80.3& 81.8 & 79.0 &  &69.2  &68.3  &70.1  \\
P2P-Net~\cite{p2p} (ICCV 2021) &   & 61.7 & 65.7 & 58.2 &  & 82.8 & 83.4 & 82.2 &  & 71.2 & 72.9 & 69.5 \\
GL~\cite{GL} (CVPR 2021) &  & 61.8 & 64.6 & 59.3 &  & 76.5 & 78.2 & 74.8 &  & 66.0 & \textbf{80.0} & 56.2 \\ 
AutoScale \cite{AutoScale} (IJCV 2022)&  & 53.7 & 57.2 & 50.7 &  & 77.3 & 78.9 & 75.8 &  & 62.0 & 67.4 & 57.4 \\ 
CLTR\cite{liang2022end} (ECCV 2022)& & - & - & - &  & 82.2 & 80.0 & 80.1 &  & 69.4 & 67.6 & 68.5\\ 
\midrule
LSC-CNN~\cite{lsc} (TPAMI 2021)& \multirow{3}{*}{Detection-Based} & 52.5 & 55.6 & 49.8 &  &  74.1&  74.6& 73.5 &  &59.3  &67.1 &53.4  \\
SDNet \cite{SDNet} (TIP 2021)&  & 56.2 & 61.3 & 52.0 &  & 78.0 & 78.9 & 77.2 &  & 63.7 & 65.1 & 62.4 \\ 
PSDNN~\cite{PSDNN} (CVPR 2019) &  & 50.2 & 53.7 & 47.1 &  &  67.0&  63.6& 70.8 &  &53.7  & 53.3 & 54.1 \\
\midrule
LSC-CNN + Crowd Hat (ours) &\multirow{3}{*}{Detection-Based}  & 57.7 & 60.4 & 55.3 &  & 80.1 & 80.4 & 79.8  &  & 70.8 & 74.3 & 67.7 \\
SDNet + Crowd Hat (ours) &  & \underline{64.3} & \underline{68.0} & \textbf{61.1} &  & \underline{83.5} & \underline{84.0} & \underline{83.1} &  & \underline{75.9} & 74.0 & \underline{77.8} \\
PSDNN + Crowd Hat (ours) &  & \textbf{65.9} & \textbf{72.6} & \underline{60.3} &  & \textbf{86.2} & \textbf{85.9} &\textbf{86.6}  &  & \textbf{78.2} & \underline{78.2} & \textbf{78.3} \\ \bottomrule
\end{tabular}}
\label{tab:localization}
\end{table*}

\begin{table}[t]
\footnotesize
\setlength{\abovecaptionskip}{0.15cm}
\setlength{\belowcaptionskip}{-0.3cm}
\centering
\caption{\textbf{Comparisons in crowd detection}, best in {\bf bold}, second in \underline{underline}.}
\setlength{\belowcaptionskip}{0cm}
\resizebox{0.474\textwidth}{!}{
\begin{tabular}{@{}lccc@{}}
\toprule
\textbf{Method}  & \textbf{Easy $\uparrow$} & \textbf{Medium $\uparrow$} & \textbf{Hard $\uparrow$} \\ \midrule
CSR-A-thr \cite{CSRNet} (CVPR 2018) &30.2 & 41.9 & 33.5 \\
LSC-CNN \cite{lsc} (TPAMI 2021) & 40.5 & 62.1 & 46.2 \\
PSDNN \cite{PSDNN} (CVPR 2019) &60.5 & 60.5 & 39.6 \\
SDNet \cite{SDNet} (TIP 2021) & 75.8 & 71.0 & 64.4 \\ \midrule
LSC-CNN + Crowd Hat (ours) &68.4  & 72.3 & 59.7 \\
PSDNN + Crowd Hat (ours)& \underline{81.5} & \textbf{78.1} & \underline{66.5} \\ 
SDNet + Crowd Hat (ours)& \textbf{84.7} & \underline{75.9} & \textbf{69.4}\\
\bottomrule
\end{tabular}
}

\label{tab:detection}
\vspace{-5mm}
\end{table}

\subsection{Implementation Details} \label{Implementation}
{The Crowd Hat module is decoupled from the detection pipeline and can be applied to a pre-trained detection model. Our framework is implemented on top of three detection-based methods, namely PSDNN~\cite{PSDNN}, LSC-CNN~\cite{lsc}, and SDNet~\cite{SDNet}. During inference, relevant data such as $t_{2d}$, $t_{1d}$ (Section \ref{sec:4.2}), and $[T_1, T_2, ..., T_{K^2}]$ (Section \ref{sec:4.4}) is saved to disk, while the detection pipeline's weights remain fixed. The Crowd Hat network (Figure \ref{fig:convolution}) is then trained using this data, with a spatial size of 64 for the 2D compressed matrix $S$ and a length of 256 intervals for the 1D distribution vector $L$. We extract local features by splitting the images into $K^2=16$ patches with a step for linear search $s$ set to 0.01. Our model is trained on 4 Nvidia RTX 3090 GPUs with a batch size of 16 for all datasets using the Adam optimizer with a learning rate of $1e-5$, over a period of 120 epochs for NWPU-Crowd dataset and 100 epoched for others. }

\subsection{Experimental Settings}
We evaluate our methods for crowd analysis tasks, including detection, localization, and counting.

\mypar{Crowd Detection} {Our method is evaluated on the WIDER-Face validation set for crowd detection using average precision (AP) as the detection metric with an IOU set to 0.5, following standard practice \cite{PSDNN,lsc,SDNet}.}

\mypar{Crowd Localization} {We evaluated our approach on three public benchmarks: JHU-Crowd++\cite{jhu}, UCF-QNRF\cite{ucf-qnrf}, and NWPU-Crowd~\cite{NWPU}. We used Precision, Recall, and F-measure as evaluation metrics. For NWPU-Crowd and JHU-Crowd++, we followed the evaluation criteria in \cite{NWPU}, which uses box labels for assessing successful matches. For UCF-QNRF, which does not have box labels, we evaluated at various distance thresholds (1 to 100 pixels) as per the standard practice in \cite{ucf-qnrf,topo}.}

\mypar{Crowd Counting} {In addition to datasets used for localization, we also performed comparisons on the ShanghaiTech dataset \cite{SH-AB}. We evaluated the counting performance using the Mean Absolute Error (MAE) and Root Mean Squared Error (RMSE) metrics, which are widely used in prior work \cite{MCNN,bys,dm,NWPU}.}

\begin{table*}[t]
\footnotesize
\setlength{\abovecaptionskip}{0.15cm}
\setlength{\belowcaptionskip}{-0.4cm}
\centering
\caption{\textbf{Ablation studies} on counting, localization, and detection. We conduct counting and localization experiments under UCF-QNRF dataset and detection experiments under WIDER-Face dataset (Hard set). We adopt PSDNN~\cite{PSDNN}, a two-stage detection based method as our baseline, which generates both proposals and bounding boxes.}
\resizebox{\textwidth}{!}{
\begin{tabular}{@{}clcccccccc@{}}
\toprule
\multirow{2}{*}{\textbf{Ablation}} & \multirow{2}{*}{\textbf{Method}} & \multicolumn{2}{c}{\textbf{Counting}} &  & \multicolumn{3}{c}{\textbf{Localization}} &  & \multicolumn{1}{c}{\textbf{Detection}} \\ \cmidrule(lr){3-4} \cmidrule(lr){6-8} \cmidrule(l){10-10} 
 &  & MAE $\downarrow$ & RMSE $\downarrow$ & \textbf{} & F1-m $\uparrow$ & Precision $\uparrow$& Recall $\uparrow$& \textbf{} & \multicolumn{1}{c}{AP $\uparrow$} \\ \midrule
\multirow{2}{*}{\textit{Ablation \uppercase\expandafter{\romannumeral1}}}  & CNN Features  &129.3  & 222.8 &  & 68.1 & 67.3 & 68.9 &  & 41.2 \\
& Output Features & \textbf{75.1} & \textbf{126.7}  &  & \textbf{86.2}  &\textbf{85.9}  &\textbf{86.6}  &  & \textbf{66.5} \\\midrule
\multirow{5}{*}{\textit{Ablation \uppercase\expandafter{\romannumeral2}}} 

& Box Area & 89.1 & 150.3 &  & 79.0 & 78.6 & 79.5 &  & 55.2 \\
 & Box Confidence & 86.9 & 147.3 &  &82.5  & 82.1 &82.9 &  &58.1  \\
 & Proposal Area & 86.7 & 145.2 &  & 83.7 &83.3 & 84.2 & & 61.9 \\
  & Proposal Confidence & 87.4 & 151.6 &  & 82.2 &80.9  & 83.5 & & 58.5 \\
 & All Output Features& \textbf{75.1} & \textbf{126.7}  &  & \textbf{86.2}  &\textbf{85.9}  &\textbf{86.6}  &  & \textbf{66.5}  \\\midrule
 \multirow{4}{*}{\textit{Ablation \uppercase\expandafter{\romannumeral3}}}
 & Baseline &137.5  & 240.1 &  & 67.0 & 63.6 & 70.8 &  & 39.6 \\ 
 & Baseline + w/o Compression & 103.6 & 171.0 &  &72.4  &71.8  &73.1  &  & 46.8   \\
 
 & Baseline + 2D & 82.1 & 138.3 &  &83.8  &83.2  &84.4  &  & 62.8   \\
 & Baseline + 2D + 1D & \textbf{75.1} & \textbf{126.7}  &  & \textbf{86.2}  &\textbf{85.9}  &\textbf{86.6}  &  & \textbf{66.5}  \\\midrule
\multirow{4}{*}{\textit{Ablation \uppercase\expandafter{\romannumeral4}}} & Baseline &137.5  & 240.1 &  & 67.0 & 63.6 & 70.8 &  & 39.6 \\
& Baseline + Region Adaptive NMS & 88.2 & 151.3 &  &84.9  &84.5  &85.3  &  & 62.2    \\
 & Baseline + Decouple-then-Align & 76.2 & 130.1 &  & 72.3 & 70.5 & 74.2 &  & 45.7 \\
  & Baseline + All & \textbf{75.1} & \textbf{126.7}  &  & \textbf{86.2}  &\textbf{85.9}  &\textbf{86.6}  &  & \textbf{66.5}  \\
 \bottomrule
\end{tabular}}

\label{tab:ablation}
\end{table*}
\subsection{Comparison to State-of-the-Art}


\vspace{3mm}
\mypar{Crowd Counting} {Table \ref{tab:counting} shows quantitative results for counting across five datasets. Density-based methods generally perform best, while existing detection methods perform worst; however, our Crowd Hat greatly improves the counting performance of detection methods, making them competitive with state-of-the-art density-based methods. Notably, our {\bf PSDNN + Crowd Hat} even outperforms some advanced density-based methods on certain datasets.}

\mypar{Crowd Localization} {Our method's evaluation under the crowd localization task is shown in Table \ref{tab:localization}. It significantly improves detection-based methods and achieves state-of-the-art performance across three datasets.}

\mypar{Crowd Detection} {Our method significantly boost the performance of detection methods on the dense face detection dataset WIDER-Face, achieving state-of-the-art detection results across three test sets, as demonstrated in Table \ref{tab:detection}.}

\begin{table}[t]
\footnotesize
\setlength{\abovecaptionskip}{0.15cm}
\centering
\caption{Comparison of the model size (M), and Inference
time (s / 100 images).}
\resizebox{0.474\textwidth}{!}{
\begin{tabular}{@{}lccc@{}}
\toprule
\textbf{Method}  & \textbf{Model Size} & \textbf{Inference Time} \\ \midrule
LSC-CNN \cite{lsc} & 35.08 & 57.73 \\
LSC-CNN + Crowd Hat (ours) &36.71 (\textcolor{red}{+4.6\%})  & 59.64 (\textcolor{red}{+3.3\%})\\\midrule
SDNet \cite{SDNet} & 40.04 & 193.54\\
SDNet + Crowd Hat (ours)& 41.67 (\textcolor{red}{+4.1\%}) & 195.47 (\textcolor{red}{+1.0\%})\\\midrule
PSDNN \cite{PSDNN} &47.51 & 14.35\\
PSDNN + Crowd Hat (ours)& 49.14 (\textcolor{red}{+3.4\%}) & 16.38 (\textcolor{red}{+14.1\%})\\ 
\bottomrule
\end{tabular}
}

\label{tab:speed}
\end{table}

\begin{figure*}[t]
  \centering
  \includegraphics[width=1\textwidth]{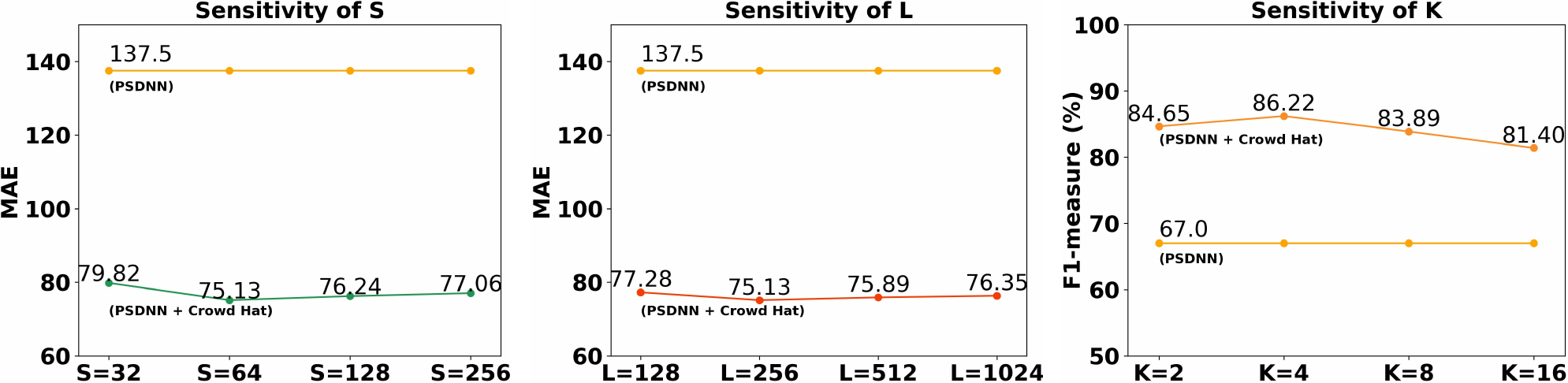}
  \caption{Studies on the sensitivity of hyperparameters on UCF-QNRF dataset.}
  \label{fig:ab}

\end{figure*}

\subsection{Ablation Studies}
Here, we study the effect of some of our key designs.

\vspace{-2mm}
\mypar{Output features vs. CNN Features} {We compared output features to CNN features (the last feature map from the backbone network) in Ablation 1 of Table \ref{tab:ablation} to study their quality. By replacing the output features with CNN features while keeping the rest constant, we found that the output features perform significantly better than CNN features.}

\mypar{Study of selected features} {In Ablation 2, we evaluated the effectiveness of each output feature used in our paper. Our results show that each output feature improves the performance of the detection baseline, and the best performance is achieved by aggregating all of these features.}

\mypar{Study of Output Feature Compression} {In Ablation 3, we evaluated the effectiveness of 2D compressed matrices and 1D distribution vectors. All other modules were included in this experiment. We started by mapping the detection outputs back to the input image without compression as ``Baseline + w/o Compression". Our results show that directly using output features without compression only provides a negligible increase in performance. However, we found that adding 2D compression matrices resulted in increased performance in all experiments. Further addition of 1D distribution vectors boosted overall performance, demonstrating the effectiveness of both 2D and 1D compression.}

\mypar{Study of Region Adaptive NMS and Decouple-then-Align Paradigm} {In Ablation 4, we evaluated the effectiveness of two important modules: Region Adaptive NMS and Decouple-then-Align Paradigm, using all output features with compression for all experiments. We found that adding Region Adaptive NMS resulted in performance increases for all tasks, particularly for Localization and Detection tasks. Decouple-then-Align Paradigm mainly boosted the performance of crowd counting task while also providing minor benefits for detection and localization tasks. The best performance was achieved by adding both modules.}

\mypar{Sensitivity of Hyperparameters} {Figure \ref{fig:ab} shows sensitivity experiments conducted on three important hyperparameters of {\bf PSDNN + Crowd Hat}: the spatial dimension of the 2D compressed matrix $S$, the spatial dimension of the 1D compressed vector $L$, and the number of regions in the region-adaptive NMS threshold $K$. Specifically, counting experiments were conducted for $S$ and $L$, while a localization experiment was conducted for $K$ using the UCF-QNRF dataset. During testing of one hyperparameter, the others were fixed to their default settings as specified in the implementation details. The performance for various values of $S$ and $L$ remained relatively stable, with the best performance achieved when $S$ = 64 and $L$ = 256. For the hyperparameter $K$, the best performance was achieved with $K=4$.}

\mypar{Running Cost Evaluations} {We compared the model size and inference time in Table~\ref{tab:speed} by processing all images into 1024 × 768 resolution and running the experiment on one RTX 3090 GPU. Our results show that after adding our Crowd Hat module, there was no significant increase in model size or inference time, indicating that our model is lightweight and can be easily adapted to different detection pipelines.}

\vspace{-1mm}
\section{Conclusion}
{In this paper, we propose a Crowd Hat module to leverage underutilized output features from bounding boxes and proposals in the detection pipeline for crowd analysis tasks. Our extensive evaluations under three different crowd analysis tasks demonstrate the effectiveness of our approach and highlight the potential of using output features as valuable assets in crowd analysis.}

\mypar{limitations and future work.} {We show that area size and confidence score have a strong correspondence with the crowd distribution, but there can be potential better features more effective. In addition, as the 1D compression is not differentiable, our model can not be trained in an end-to-end manner.
}

\nocite{*}
{\small
\bibliographystyle{ieee_fullname}
\bibliography{main}
}
\clearpage
\appendix
\section{Detailed Evaluation Metrics}
\paragraph{Crowd Counting}
Mean Absolute Error (MAE) and Root Mean Squared Error (RMSE) are widely used as counting metrics, and they are defined as follows:

\begin{equation}
    MAE = \frac{1}{N}\sum_{i=1}^{n} \left | e_{i} - gt_{i}    \right | 
\end{equation}

\vspace{-0.05cm}

\begin{equation}
    RMSE = \sqrt{\frac{1}{N}\sum_{i=1}^{n} (e_{i} - gt_{i} )^{2} } 
\end{equation}
Here, $e_{i}$ and $gt_{i}$ represent the estimated count and ground truth count of crowds, respectively, and N is the total number of images.

\paragraph{Crowd Localization}
F1 Measure, Precision, and Recall are commonly used as metrics for crowd localization, as proposed in \cite{NWPU}. We denote the two point sets of prediction results as $P_p$ and ground truth as $P_g$, and construct a Bipartite Graph $G_{p,s}$ for the two sets. Then, we compute the distance matrix of $P_p$ and $P_g$. If the distance between $p_p \in P_p$ and $p_g \in P_g$ is less than a predefined distance threshold $\sigma$, we consider $p_p$ and $p_g$ to be successfully matched, and obtain a boolean match matrix (True and False denote matched and non-matched) corresponding to each element of the distance matrix. Finally, by implementing the Hungarian algorithm, we obtain a Maximum Bipartite Matching for $G_{p,s}$. Based on the counts of True Positive (TP), False Positive (FP), and False Negative (FN), we can compute Precision (P), Recall (R), and F1 Measure ($F_1$) as follows:
 
 \begin{equation}
     P=\frac{TP}{TP+FP},R=\frac{TP}{TP+FN},F_{1} =\frac{2PR}{P+R}
 \end{equation}

\paragraph{Crowd Detection}\
Following standard practices, we adopt the average precision (AP) as the detection metric, with an Intersection over Union (IOU) threshold of 0.5.

\section{Datasets}
\noindent
{\bf WIDER-Face} \cite{wider-face} is a dense face detection dataset consisting of 32,203 images and 393,703 face labels, which exhibit high variability in terms of scale, pose, and occlusion.

\noindent
{\bf ShanghaiTech} \cite{SH-AB}  consists of two independent subsets, Part A and Part B. Part A contains highly congested images collected from the Internet, while Part B is comprised of images taken from the busy streets of metropolitan areas in Shanghai.

\noindent
{\bf UCF-QNRF} \cite{ucf-qnrf} contains 1535 images, which exhibit a much wider range of crowd counts compared to the previous datasets, making it a more challenging dataset for crowd analysis

\noindent
{\bf JHU-Crowd++} \cite{jhu} is a large-scale unconstrained dataset that comprises a total of 4,372 images, containing 1,515,005 head annotations and captured under a variety of conditions. The dataset includes challenging images captured under various weather-based degradation, as well as some negative samples that may be detected as false positives.

\noindent
{\bf NWPU-Crowd} \cite{NWPU} is the largest crowd analysis dataset, consisting of 5,109 images and 2,133,375 annotated heads with varying crowd densities. For an authentic evaluation of crowd counting and localization, we report our results from the official website of NWPU-crowd.

\begin{figure*}[t!]
  \centering
  \begin{subfigure}{\linewidth}
    \includegraphics[width=1\linewidth]{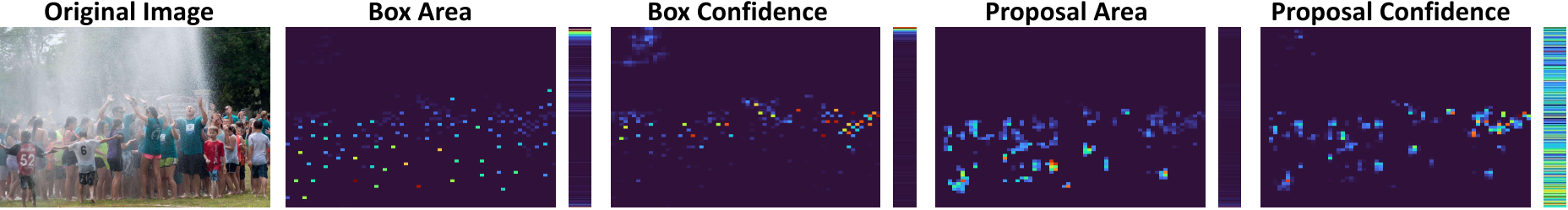}
    \caption*{}
    \vspace{-0.45cm}
  \end{subfigure}
  \begin{subfigure}{\linewidth}
    \includegraphics[width=1\linewidth]{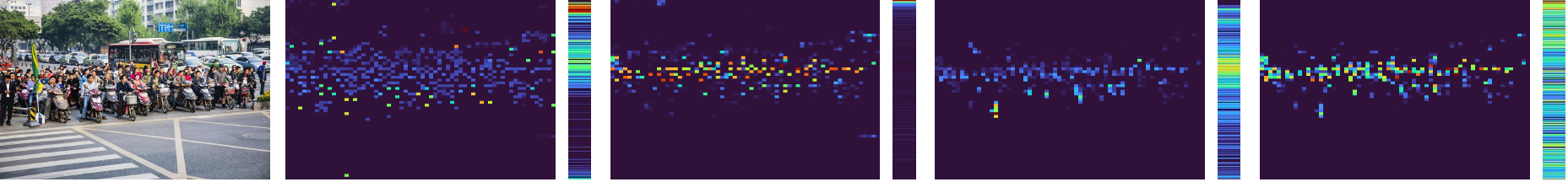}
    \caption*{}
    \vspace{-0.45cm}
  \end{subfigure}
  \begin{subfigure}{\linewidth}
    \includegraphics[width=1\linewidth]{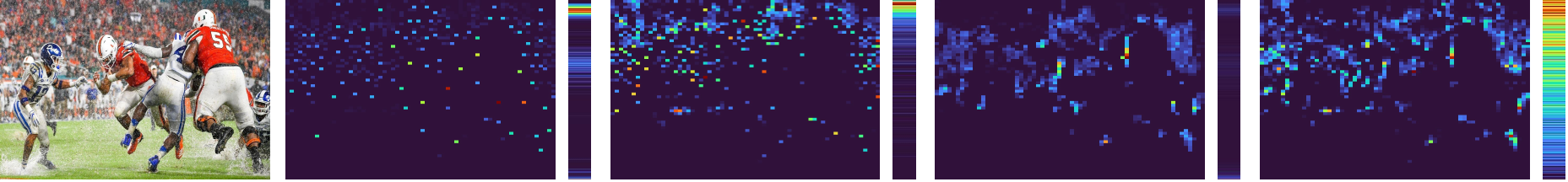}
    \caption*{}
    \vspace{-0.45cm}
  \end{subfigure}
  \begin{subfigure}{\linewidth}
    \includegraphics[width=1\linewidth]{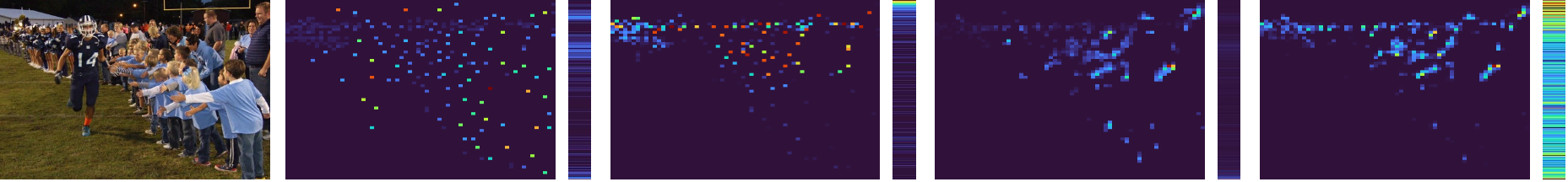}
    \caption*{}
    \vspace{-0.45cm}
  \end{subfigure}
  \begin{subfigure}{\linewidth}
    \includegraphics[width=1\linewidth]{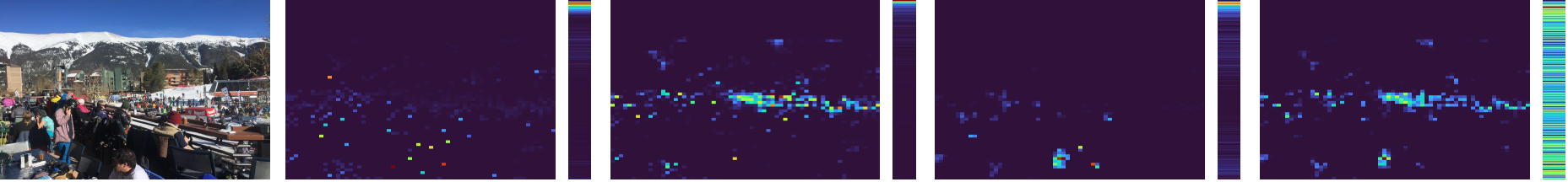}
    \caption*{}
    \vspace{-0.45cm}
  \end{subfigure}
  \begin{subfigure}{\linewidth}
    \includegraphics[width=1\linewidth]{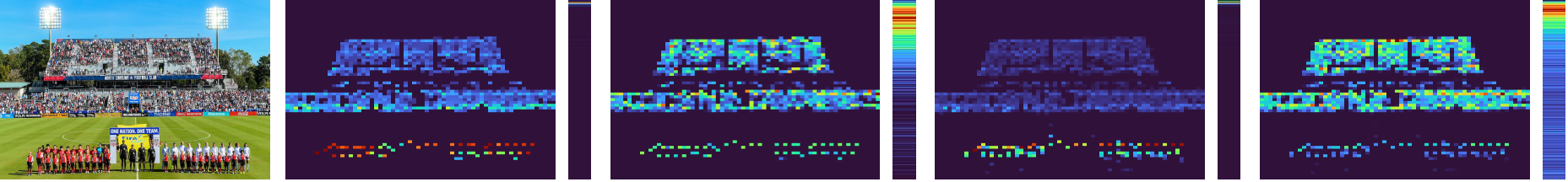}
    \caption*{}
    \vspace{-0.45cm}
  \end{subfigure}
  \begin{subfigure}{\linewidth}
    \includegraphics[width=1\linewidth]{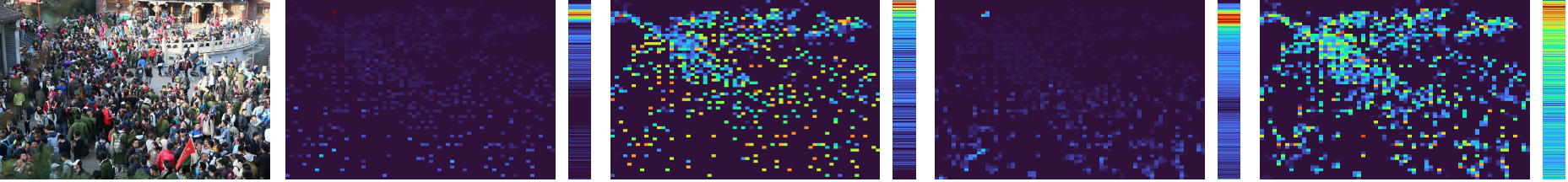}
    \caption*{}
    \vspace{-0.45cm}
  \end{subfigure}
  \begin{subfigure}{\linewidth}
    \includegraphics[width=1\linewidth]{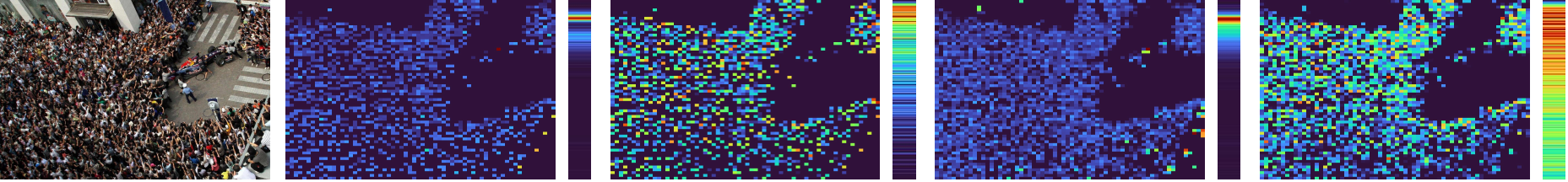}
    \caption*{}
    \vspace{-0.45cm}
  \end{subfigure}
  \centering
  \caption{To visualize the 2D-1D feature compression, we present both the 2D compression matrices (on the left) and 1D distribution vectors (to their right) for each output feature. In the 1D distribution vectors, we denote 0 at the top and 1 at the bottom. Additionally, we include the original image in the leftmost column for reference. \emph{Zoom in for better visualization.}}
  \label{fig:compress}
\end{figure*}

\section{Visualization of Compression}
Figure \ref{fig:compress} provides additional visualization of the 2D-1D feature compression achieved by the {\bf PSDNN + Crowd Hat}. Heat map is adopted where 2D compression is on the left and 1D compression is on the right.

\section{Visualization of Detection}
\begin{figure*}[t!]
  \centering
  \begin{subfigure}{\linewidth}
    \includegraphics[width=1\linewidth]{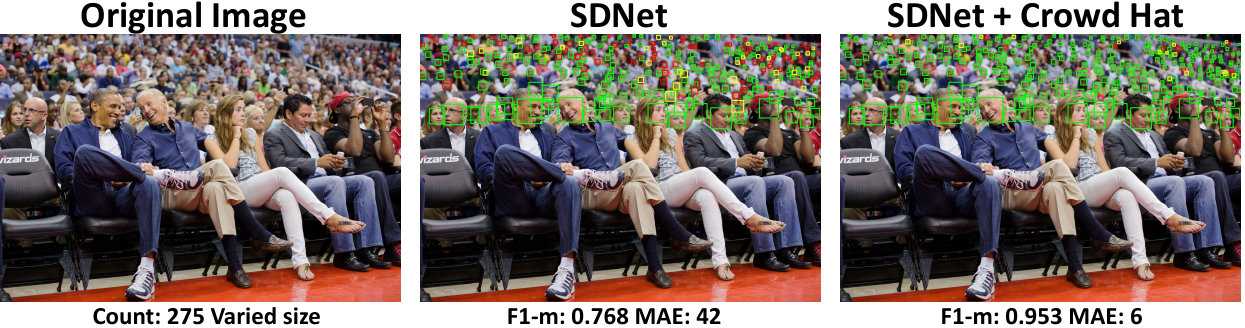}
    \caption*{}
    \vspace{-0.3cm}
  \end{subfigure}
  \begin{subfigure}{\linewidth}
    \includegraphics[width=1\linewidth]{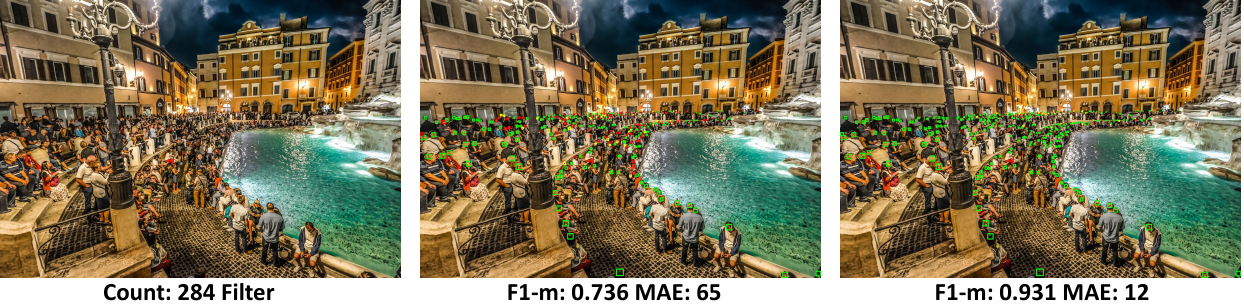}
    \caption*{}
    \vspace{-0.3cm}
  \end{subfigure}
  \begin{subfigure}{\linewidth}
    \includegraphics[width=1\linewidth]{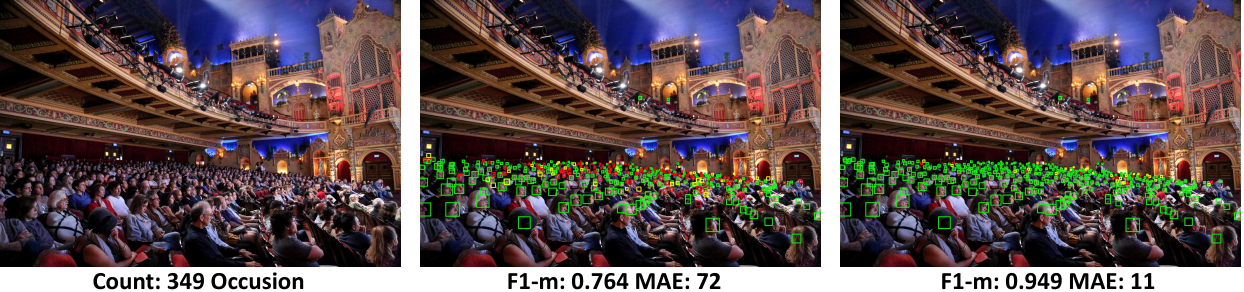}
    \caption*{}
    \vspace{-0.3cm}
  \end{subfigure}
  \begin{subfigure}{\linewidth}
    \includegraphics[width=1\linewidth]{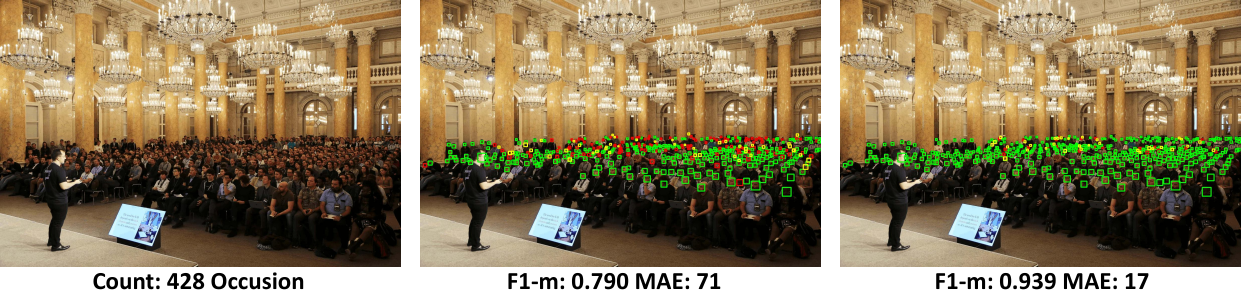}
    \caption*{}
    \vspace{-0.3cm}
  \end{subfigure}
  \begin{subfigure}{\linewidth}
    \includegraphics[width=1\linewidth]{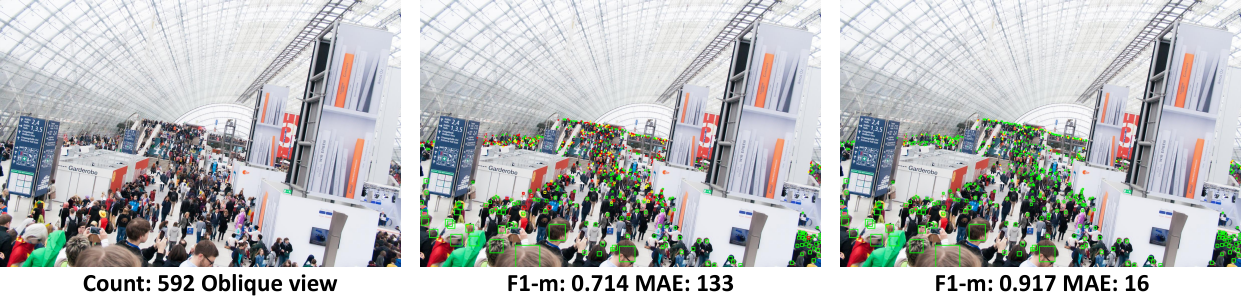}
    \caption*{}
    \vspace{-0.3cm}
  \end{subfigure}
  \centering
  \caption{Visualization of Detection in Low Density Crowd.}
  \label{fig:detection1}
\end{figure*}

\begin{figure*}[t!]
  \centering
  \begin{subfigure}{\linewidth}
    \includegraphics[width=1\linewidth]{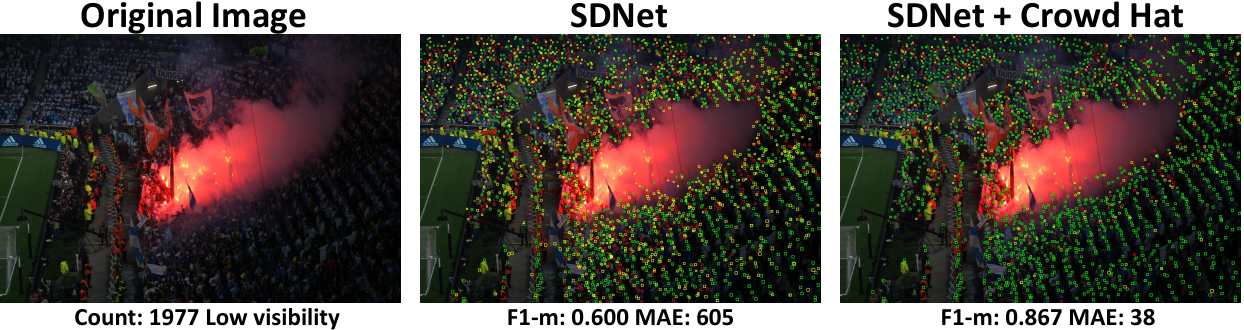}
    \caption*{}
    \vspace{-0.3cm}
  \end{subfigure}
  \begin{subfigure}{\linewidth}
    \includegraphics[width=1\linewidth]{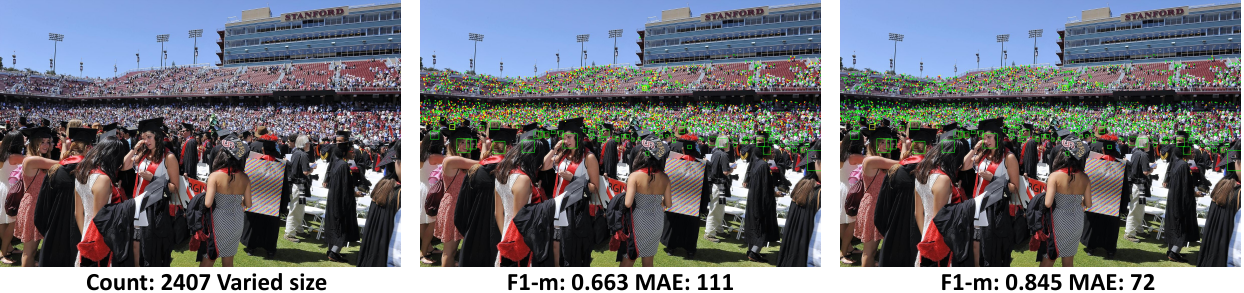}
    \caption*{}
    \vspace{-0.3cm}
  \end{subfigure}
  \begin{subfigure}{\linewidth}
    \includegraphics[width=1\linewidth]{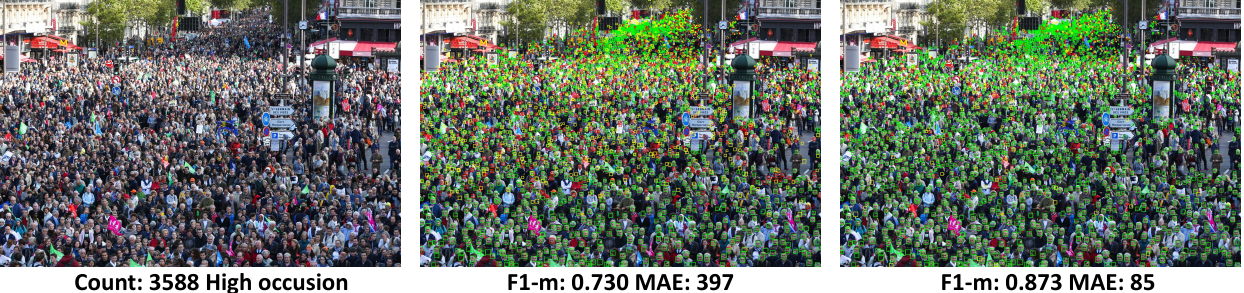}
    \caption*{}
    \vspace{-0.3cm}
  \end{subfigure}
  \begin{subfigure}{\linewidth}
    \includegraphics[width=1\linewidth]{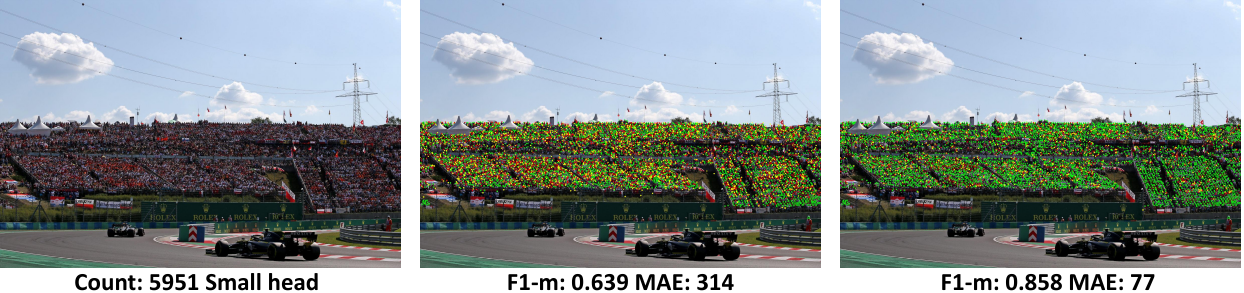}
    \caption*{}
    \vspace{-0.3cm}
  \end{subfigure}
  \begin{subfigure}{\linewidth}
    \includegraphics[width=1\linewidth]{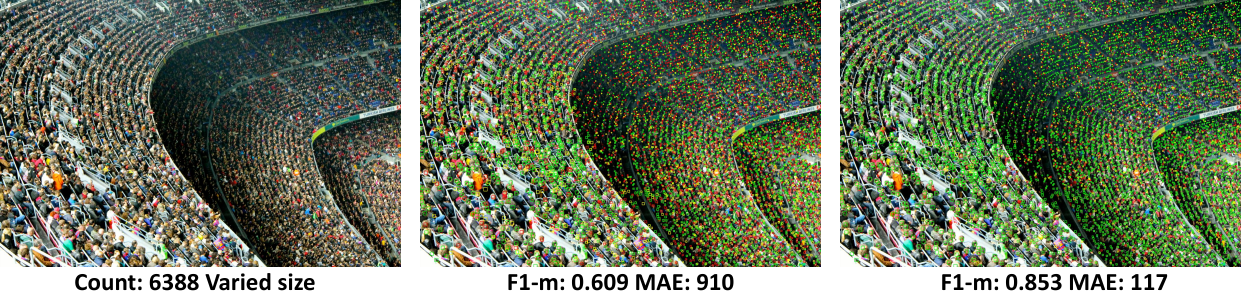}
    \caption*{}
    \vspace{-0.3cm}
  \end{subfigure}
  \centering
  \caption{Visualization of Detection in High Density Crowd.}
  \label{fig:detection2}
\end{figure*}

Figures \ref{fig:detection1} and \ref{fig:detection2} depict the detection results of SDNet with and without our proposed Crowd Hat module. Following the practice in \cite{NWPU,PSDNN}, we use green boxes to indicate true positives based on ground truth annotations, red boxes for false negatives, and yellow boxes for false positives, for better clarity.

\clearpage
\end{document}